\documentclass[10pt,twocolumn,letterpaper]{article}

\usepackage{iccv}              
\usepackage{float}
\usepackage{amsmath}
\usepackage{algorithm}
\usepackage{algpseudocode}
\usepackage{amssymb}
\usepackage{booktabs}
\usepackage{multirow}
\usepackage{graphicx}
\usepackage{pifont} 
\usepackage{amssymb}
\usepackage{mathabx}
\usepackage{booktabs}
\usepackage{array}
\usepackage[margin=1in]{geometry}

\usepackage{graphicx}
\usepackage{array}
\usepackage{booktabs}
\usepackage{caption}
\usepackage[accsupp]{axessibility}

\definecolor{iccvblue}{rgb}{0.21,0.49,0.74}
\usepackage[pagebackref,breaklinks,colorlinks,allcolors=iccvblue]{hyperref}

\def\paperID{13541} 
\def\confName{ICCV}
\def\confYear{2025}

\title{\textit{SemGes}: Semantics-aware Co-Speech Gesture Generation using Semantic Coherence and Relevance Learning}

\author{
Lanmiao Liu$^{1,2,3}$\; Esam Ghaleb$^{1,2}$\; Asl\i~Özy\"{u}rek$^{1,2}$ and Zerrin Yumak$^{3}$\\
\normalsize$^{1}$Max Planck Institute for Psycholinguistics\;
\normalsize$^{2}$Donders Institute for Brain Cognition and Behaviour\;
\normalsize$^{3}$Utrecht University\\ 
{\tt\small \{lanmiao.liu, esam.ghaleb, asli.ozyurek\}@mpi.nl \;\;z.yumak@uu.nl}
}

\begin{document}
\maketitle
\begin{abstract}
Creating a virtual avatar with semantically coherent gestures that are aligned with speech is a challenging task. Existing gesture generation research mainly focused on generating rhythmic beat gestures, neglecting the semantic context of the gestures. In this paper, we propose a novel approach for semantic grounding in co-speech gesture generation that integrates semantic information at both fine-grained and global levels. Our approach starts with learning the motion prior through a vector-quantized variational autoencoder. Built on this model, a second-stage module is applied to automatically generate gestures from speech, text-based semantics and speaker identity that ensures consistency between the semantic relevance of generated gestures and co-occurring speech semantics through semantic coherence and relevance modules. Experimental results demonstrate that our approach enhances the realism and coherence of semantic gestures. Extensive experiments and user studies show that our method outperforms state-of-the-art approaches across two benchmarks in co-speech gesture generation in both objective and subjective metrics. The qualitative results of our model, code, dataset and pre-trained models can be viewed at \href{https://semgesture.github.io/}{https://semgesture.github.io/}.
\end{abstract}    
\section{Introduction}
\label{sec:intro}

\begin{figure}[htbp]
    \centering\includegraphics[width=1.0\linewidth]{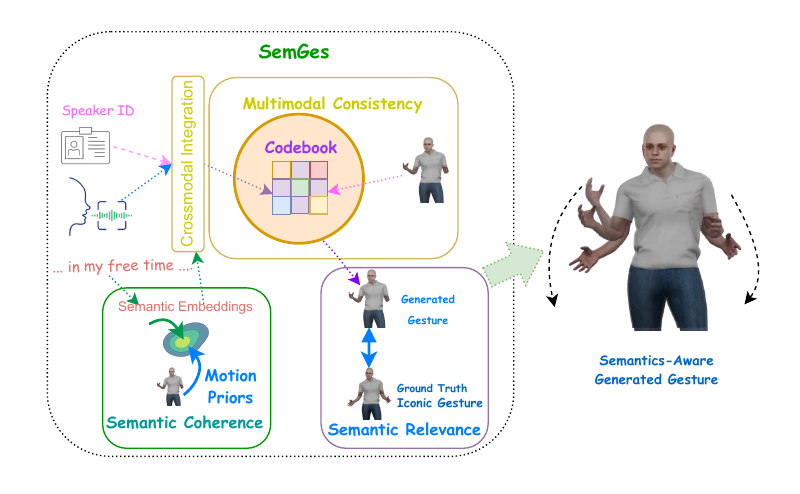}
  \caption{\emph{SemGes} integrates audio, text-based semantics, and speaker identity to produce both contextually relevant (discourse-level) and fine-grained (local) gestures. A semantic coherence module aligns text and motion embeddings. The multimodal consistency loss synchronizes the quantized multimodal representations to match the quantized learned motion features for final speech-driven semantics-aware gesture generation. The semantic relevance loss selectively emphasizes gestures with semantic annotations.}
  \label{fig:firstpage} 
   \vspace{-0.2cm}
\end{figure}

Human language is inherently multimodal, with gestures and speech complementing each other to convey pragmatic and semantic information \cite{mcneill1992hand, kendon2004gesture}.
Co-speech gestures are non-verbal cues that are uniquely related to co-occurring speech, pragmatically, semantically, and temporally.
For example, \emph{representational iconic gestures} that visually express the semantic content of speech and interact with spoken language~\cite{kendon2004gesture, HollerLevinson2019, ozyurek2014hearing, ghaleb-etal-acl-2025, ghaleb2024learning}.
A long-standing goal in Computer Vision is to create digital humans that use non-verbal cues in sync with speech. Gesture generation—synthesizing movements from co-occurring speech, masked motion, or speaker identity—has advanced to enhance AI agents’ expressiveness and realism~\cite{liu2023survey}. However, much of the focus has gone into generating rhythmic beat gestures with limited semantic information, leaving representational gestures that convey semantic messages (e.g., iconic) less explored \cite{liu2023survey, nyatsanga2023comprehensive}.

Generating spontaneous and semantically rich gestures from speech comes with multiple challenges. First, it requires capturing global discourse-level information and local fine-grained details (e.g., salient words) to generate speech-driven gestures that reflect the intended meaning and align with speech temporally and semantically. 
Second, existing methods often generate repetitive and short sequences that do not span the full range of expressive motions required for natural communication.
To leverage semantics when generating gestures, researchers have attempted to align motion with speech representations at a global level, e.g., by leveraging pre-trained semantic representations such as CLIP~\cite{zhi2023livelyspeaker} or focusing on semantically important keywords~\cite{ao2023gesturediffuclip,zhang2024semantic}.  Nonetheless, they often fail to (i) unify global and local semantic modelling within a single framework and  (ii) exploit the relevance of the semantic information in guiding gesture generation \cite{liu2022beat}. At the same time, raw audio features and speaker identity are relevant to the timing and style of gestures.
In this paper, we address these limitations by integrating speech, speech semantics, and gesturing style, exploiting semantic information at different levels.

Specifically, we propose a two-stage framework, namely, SemGes, that integrates speech, text-based semantics, and speaker identity into a unified gesture-generation model (see Figure~\ref{fig:firstpage}). In stage 1, we build motion prior of holistic gestures (\ie, body and hands) by training a vector-quantized variational autoencoder (VQ-VAE) to learn an efficient, compositional motion latent space.
This stage results in a robust motion encoder \& decoder and quantized codebooks that can reconstruct naturalistic gestures while allowing the reuse of learned codebook entries. Stage 2 leverages the learned motion priors to drive gesture synthesis by fusing three modalities using a cross-modal Transformer encoder: (i) text-based semantics, (ii) raw-audio speech features, and (iii) speaker identity for style consistency. 
We impose a \emph{semantic coherence} loss that aligns text-based embeddings with the VQ-VAE motion latent space and a \emph{semantic relevance} loss that emphasises representational gestures (\eg iconic and metaphoric gestures). A multimodal consistency objective ensures the fused multimodal representations are compatible with the learned motion codebooks, enabling the generation of gestures that are both semantically rich and visually natural. 
Finally, we introduce a simple but effective long-sequence inference strategy that smoothly combines overlapping motion clips for extended durations. To summarize our contributions,
\begin{itemize}
    \item We introduce a novel framework, \emph{SemGes}, that first learns a robust VQ-VAE motion prior for body and hand gestures, and then generates gestures driven by fused speech audio, text-based semantics, and speaker identity in a cross-modal transformer.
    \item Our method jointly captures discourse-level context via a semantic coherence loss and fine-grained representational gestures (e.g., iconic, metaphoric) via a semantic relevance loss.
    
    \item  We propose an overlap-and-combine inference algorithm that maintains smooth continuity over extended durations. 
    
    \item Extensive experiments on two benchmarks, namely, the BEAT \cite{liu2022beat} and TED Expressive \cite{liu2022learning} datasets show that our method outperforms recent baselines in both objective metrics (e.g., Fr\'echet Gesture Distance (FGD), diversity, semantic alignment) and user judgment of generated gestures. 
\end{itemize}
\section{Related Work}
\label{sec:RelatedWork}

\paragraph{Data-driven Co-Speech Gesture Generation.} 
Current gesture generation approaches are based on generative deep neural networks.
These approaches use advanced models such as Transformers \cite{liu2023emage}, Generative Adversarial Networks~\cite{li2022danceformer}, Normalizing Flows~\cite{henter2020moglow,liu2023human}, Vector Quantized Variational Autoencoder(VQ-VAE)~\cite{guo2022tm2t} and Denoising Diffusion Probabilistic Models~\cite{tevet2022human}. In addition, researchers have explored the impact of different model inputs on the naturalness and appropriateness of generated gestures.
Various modal inputs have been used, such as text~\cite{guo2022generating}, audio~\cite{zhu2023taming,yang2023diffusestylegesture}, image~\cite{liu2022audio,ning2024dctdiff}, and speaking style~\cite{alexanderson2020style}. For a comprehensive survey, we refer to \citet{nyatsanga2023comprehensive}. 
Although there have been significant improvements in this field, current methods fall short in generating semantically grounded gestures at a fine-grained level. In other words, while the generated motions look convincing at first glance, they do not match well with the meaning of the text, or they mostly focus on beat-type gestures.

\paragraph{Semantics-aware Co-Speech Gesture Generation.}
A group of work focused on semantics-aware gesture generation where the semantic information is handled in two ways: global semantics and local semantics. Methods that focus on global semantic information \cite{zhi2023livelyspeaker,chen2024diffsheg,kucherenko2020gesticulator} align gestures with text or audio, but they fall short in generating gesture types matching the semantic context, such as iconic, metaphoric and deictic gestures. To capture a wider range of semantic gestures, works like \cite{ao2022rhythmic,ao2023gesturediffuclip,liu2022beat} adopt local semantic-aware modelling by integrating the semantic salient words to the neural network. However, these approaches often fail to ensure that the generated gestures align with both the broader audio or textual context and a combination of global and local semantics. \citet{voss2023augmented,liang2022seeg} incorporate both global and local semantics, however, they require extensive annotations. Recently, \citet{zhang2024semantic} employed a generative retrieval framework based on LLMs to address the sparsity problem in datasets with semantic gestures. However, they do not explicitly model the different types of gestures~\cite{liu2022beat} or gesture phases \cite{ferstl2020adversarial} grounded in linguistic research. Moreover, there is still not enough understanding of the impact of different annotations and fine-grained semantics.

Substantial research \cite{zhi2023livelyspeaker,ao2023gesturediffuclip,zhang2024semantic,liu2023emage} focused on two-stage latent space generative modelling to overcome the limitations of co-speech gesture generation and to generate more naturalistic and diverse gestures. These approaches first learn a latent space and then model gestures probabilistically, effectively integrating the strengths of different methods in different stages. \citet{zhang2024semantic,liu2023emage} capture complex dependencies in the latent space using VQ-VAE while \citet{zhi2023livelyspeaker} employs CLIP~\cite{tevet2022motionclip,zhang2024kinmo} to align text and motion embeddings. \citet{ao2023gesturediffuclip} introduces a diffusion-based model that leverages semantic awareness, while \citet{liu2023emage} utilizes a transformer-based approach to generate holistic body gestures. In contrast with two-stage generative modeling approaches, end-to-end methods such as \cite{zhu2023taming, ao2022rhythmic}, are prone to jittering artifacts, especially in hand-motion generation.

Our model contributes to the research line on semantics-aware co-speech gesture generation by taking into account both global and local semantics. Inspired by the previous work, we employ a two-stage latent space generative modelling for high-quality motion representation. We learn semantic coherence between text and gestures globally with cosine similarity. Moreover, our model takes into account the semantic relevancy of gesture types with minimally required annotations. In contrast with other semantic learning models, we focus on annotations with different gesture types embedded in linguistic research. Our work is closest to CAMN \cite{liu2022beat} in that sense; however, CAMN does not include semantic coherence learning by aligning text and gestures' latent space globally.

\section{Methodology}
\label{sec:Methodology}
We propose a two-stage approach that generates co-speech gestures by grounding them in raw speech, text-based semantics, and speaker identity. In Section \ref{sec:VQ-VAE}, we introduce a VQ-VAE encoder-decoder that learns a robust motion prior. Section \ref{sec:stage_2} details our gesture synthesis and inference pipeline based on speech, semantics, and identity.
\begin{figure}[t!]
  \centering
  \includegraphics[width=1.0\linewidth]{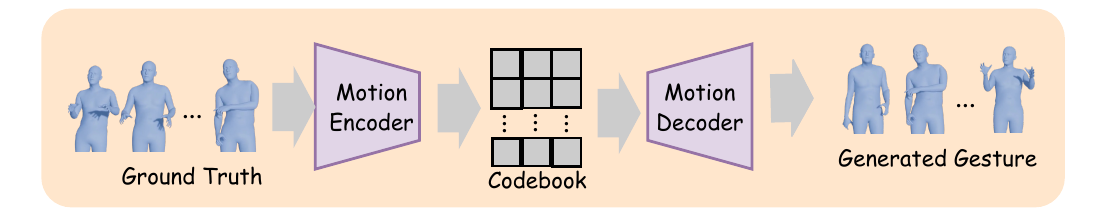}
  \caption{We pre-train two VQ-VAEs by reconstructing body and hand motions with a dedicated codebook for each.}
  \label{fig:1}
  \vspace{-0.2cm}
\end{figure}
\begin{figure*}[htbp]
  \centering
  \includegraphics[width=0.99\linewidth]{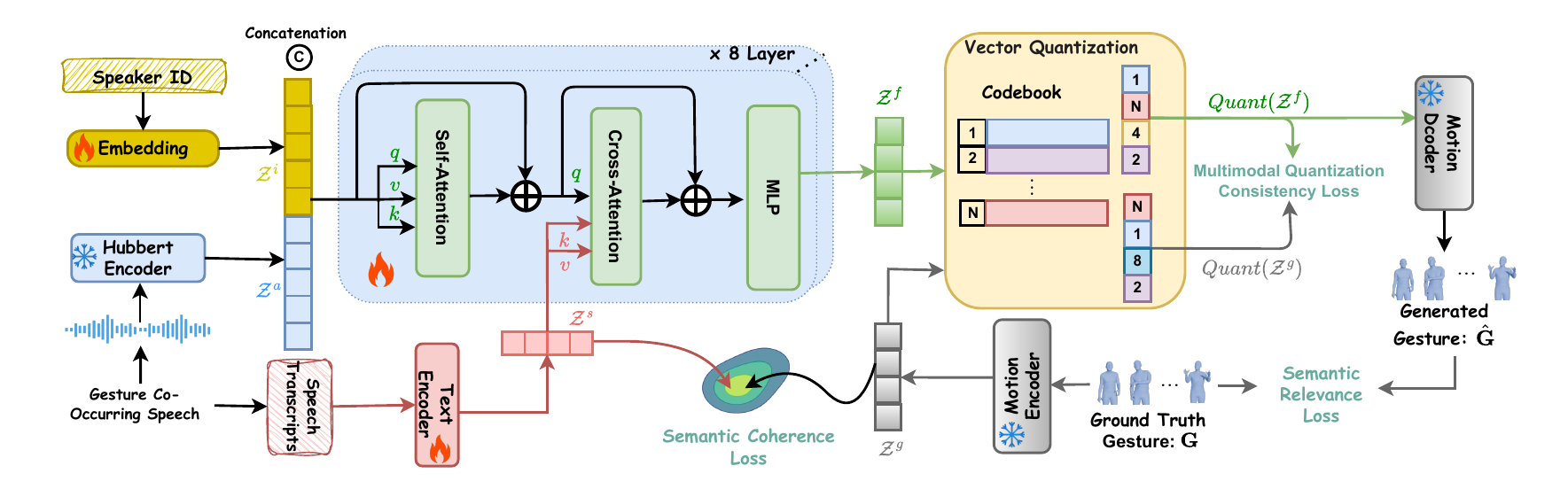}
  \caption{SemGes employs three training pathways: (1) Global semantic coherence, which minimizes latent disparities between gesture and text encoders; (2) Multimodal Quantization learning, where integrated multimodal representation codes are aligned with quantized motion to decode them into hand and body movements; and (3) Semantic relevance learning, which emphasizes semantic gestures.}
  \label{fig:2}
  \vspace{-0.2cm}
\end{figure*}
\paragraph{Problem formulation.}
Our goal is to generate hand gestures 
\(\boldsymbol{G}^{h} = (g^{h}_1, \dots, g^{h}_T) \in \mathbb{R}^{T \times J}\) 
and body gestures
\(\boldsymbol{G}^{b} = (g^{b}_1, \dots, g^{b}_T) \in \mathbb{R}^{T \times J}\),
where \(T\) is the number of time steps and \(J\) the number of joints (e.g., 38 for hands, 9 for body). Each motion vector \(g^{h}_t\) or \(g^{b}_t\) is encoded in a Rot6D representation, capturing joint rotations at time \(t\). 

To model human motion of body and hands, we first learn a motion generator \(\mathcal{M}_{g}\) (Stage 1), which synthesizes a plausible motion sequence:
\begin{equation}
\label{eq:1}
\arg\min_{\mathcal{M}_{g}}
\Bigl\|
\boldsymbol{G}
-\mathcal{M}_{g}(g_1, \dots, g_T)
\Bigr\|.
\end{equation}

\noindent
Next, we condition on (i) the raw input ${a}$udio \(\boldsymbol{A} = (a_1, \dots, a_T)\), 
(ii) the speaker identity embedding \(I\), and 
(iii) the text-based semantic embeddings of the speech \(\boldsymbol{S} = (s_1, \dots, s_T)\).
Our second-stage model \(\mathcal{M}_{{a,s, i}}\) uses these inputs to generate a latent sequence that the motion generator \(\mathcal{M}_{g}\) then decodes into naturalistic gestures:

\begin{equation}
\scriptsize
\label{eq:2}
\arg\min_{\mathcal{M}_{a,s,i}}
\Bigl\|
\boldsymbol{G}
- \mathcal{M}_{g}\bigl(\mathcal{M}_{a,s,i}(\boldsymbol{A}, \boldsymbol{S}, I)\bigr)
\Bigr\|.
\end{equation}

\subsection{Stage 1: Learning Efficient Codebooks \& Compositional Motion Priors}
\label{sec:VQ-VAE}
Realistic co-speech gestures require modelling the sequential motion of both body and hand joints. Rather than learning a single representation for the entire body, we adopt a compositional approach, using a discrete codebook of learned representations specific to each part (hands \& body). 
Any gesture motion can then be represented by selecting appropriate codebook entries. Following \cite{van2017neural, yi2022generating, liu2023emage}, we employ a VQ-VAE architecture (see Fig.~\ref{fig:1}) with encoder \(\mathcal{E}_{m}\) and decoder \(\mathcal{D}_{m}\). Given hand motion 
\(\boldsymbol{G}^{h} \in \mathbb{R}^{T\times J}\) 
and body motion 
\(\boldsymbol{G}^{b} \in \mathbb{R}^{T\times J}\),
the encoder produces latent vectors \(\hat{z}^h\) and \(\hat{z}^b\), which are quantized by selecting the nearest entries in the codebooks. Formally,

\begin{equation}
\label{eq:3}
\mathbf{q}(\boldsymbol{\hat{z}})
=  \arg \min_{z^i \in \mathcal{Z}} 
\bigl\| \hat{z}^j - z^i \bigr\|,
\end{equation}
where \(z^i\) are the learned codebook entries, and \(\hat{z}^j\) denotes an element of the latent vector for either hand or body. We train the VQ-VAE via a straight-through gradient estimator, minimizing:

\begin{equation}
\scriptsize
\label{eq:4}
\begin{split}
\mathcal{L}_{\text{VQ-VAE}}
=
&\ \bigl\|\mathbf{g}- \hat{\mathbf{g}} \bigr\|^2
 + \bigl\|\dot{\mathbf{g}}- \hat{\dot{\mathbf{g}}}\bigr\|^2
 + \bigl\|\ddot{\mathbf{g}}- \hat{\ddot{\mathbf{g}}}\bigr\|^2 \\
&\ + \bigl\| \text{sg}[\mathbf{E(g)}] - \mathbf{q}(\boldsymbol{\hat{z}})\bigr\|^2 
 + \bigl\|\mathbf{E(g)} - \text{sg}[\mathbf{q}(\boldsymbol{\hat{z}})]\bigr\|^2,
\end{split}
\end{equation}
where the first three terms reconstruct joint positions, velocities, and accelerations, and the last two terms implement the VQ-VAE commitment loss \cite{van2017neural}. 

By the end of this stage, we have motion ($m$) encoder ($\mathcal{E}_{m}$), decoder ($\mathcal{D}_{m}$) and codebooks (\( Quant^m(\cdot) \)) for hands and body. In the next section (Section~\ref{sec:stage_2}), we show how this discretized motion of hands and body guides speech, semantics and speaker identity-driven generation to produce realized co-speech gestures.

\subsection{Stage 2: Speech and Identity Driven Semantic Gesture Generator}
\label{sec:stage_2}
This stage focuses on generating gestures conditioned on three inputs: speech embeddings, text-based semantic embeddings, and speaker identity. As illustrated in Figure~\ref{fig:2}, the second-stage architecture has three main modules, which we elaborate on in the following subsections.

\begin{figure}[htbp]
  \centering
  \includegraphics[width=0.8\linewidth]{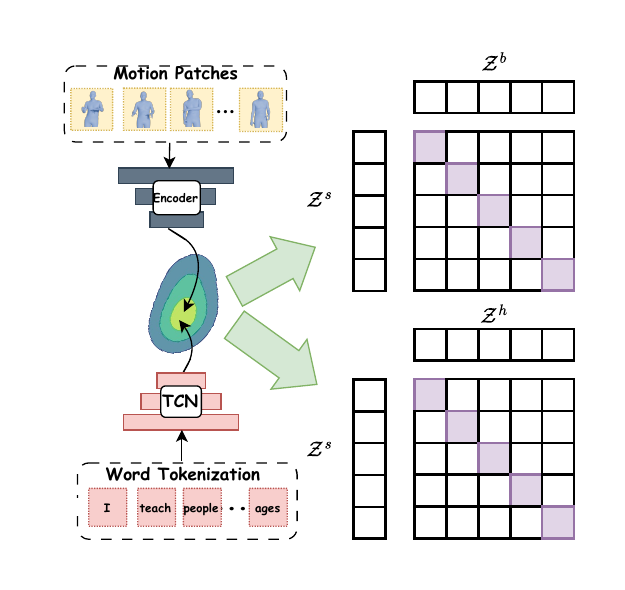}
  \caption{Semantic Coherence Embedding Learning.}
  \label{fig:3} 
  \vspace{-0.2cm}
\end{figure}

\subsubsection{Semantic Coherence Embedding Learning}
\label{subsec:semantic_coherence}

To align text-based semantics with motion embeddings, we introduce a shared embedding space for both motion priors and speech transcripts. Specifically, we embed word tokens using a pre-trained FastText model~\cite{bojanowski2017enriching}, then feed these embeddings into a trainable text-based semantic encoder $\mathcal{E}_{s}$.
At the same time, we use the \emph{pre-trained motion encoder} $\mathcal{E}_{m}$ from Stage\,1 to encode ground-truth gesture sequences. Thus, for a batch of paired (gesture, transcript) samples, we get:
\begin{equation}
\scriptsize
\label{eq:text_feat}
\mathcal{Z}^{S} = \mathcal{E}_{s}(S), 
\quad
\mathcal{Z}^{h} = \mathcal{E}^{h}_{m}(G^h),
\quad
\mathcal{Z}^{b} = \mathcal{E}^{b}_{m}(G^b),
\end{equation}
where $S$ is the tokenized speech transcript, and $G^h$ and $G^b$ correspond to the ground-truth hand gesture sequence and body gesture sequence, respectively. \( \mathcal{Z}^{h}  \) and \( \mathcal{Z}^{b} \) represent the hand and body ground-truth motion encodings from Stage 1, and \( \mathcal{Z}^{s} \) denotes the text-based semantic encoder output.  

\paragraph{Semantic Coherence Loss.}
We maximize the similarity of correct (gesture, transcript) pairs and minimize it for mismatched pairs, enforcing \emph{semantic coherence}. This aligns gestures and textual semantics in a common space while keeping $\mathcal{E}_{m}$ frozen, as illustrated in Figure~\ref{fig:3}. We impose the semantic coherence constraint separately on both hand and body movements to align gestures with transcripts in the shared embedding space. Specifically, we introduce two distinct cosine similarity losses: one between the text encoder output and the hand motion latent encoding and another between the text encoder output and the body motion latent encoding. Formally, we minimize:

\begin{equation}
\scriptsize
\mathcal{L}_{\text{semantic-coherence}} = 1 - \cos(\mathcal{Z}^{h}, \mathcal{Z}^{s}) + 1 - \cos(\mathcal{Z}^{b}, \mathcal{Z}^{s}),
\end{equation}
where the function \( \cos(\cdot, \cdot) \) measures cosine similarity. 

\subsubsection{Crossmodal Integration}
\label{subsec:cross_modal_integration}
SemGes supports multi-modal inputs in the second training stage by combining audio features and speaker identity with semantic text embeddings using a Transformer encoder with self and cross-attention layers (see Figure~\ref{fig:2}). We begin by extracting HuBERT features~\cite{hsu2021hubert} from raw speech (keeping the HuBERT encoder frozen). We concatenate audio features $\mathcal{Z}^{a}$ and speaker embeddings $\mathcal{Z}^{i}$, resuling in $\mathcal{Z}^{r}$, which we feed into a \emph{self-attention layer}.

Next, we use a \emph{cross-attention layer} that takes $\mathcal{Z}^{r}$ as the query and the motion-aligned text-based semantic features $\mathcal{Z}^{s}$ as the key-value pair.
The final hidden representation $\mathcal{Z}^{f}$ serves as the \emph{multimodal latent code} that drives gesture synthesis when passed to our vector quantization and VQ-VAE-based motion decoder, which is learned in our first stage (see the yellow box in Figure~\ref{fig:2}).

\paragraph{Multimodal Quantization Consistency Loss.}
SemGes quantizes the multimodal latent code using separate hand and body codebooks. To align this code with the ground-truth motion latent codes, we apply independent quantization consistency losses for each component. Specifically, the quantization loss is defined as:
\begin{equation}
\scriptsize
\begin{split}
\mathcal{L}_{\text{quantization}} = \left\| Quant^h(\mathcal{Z}^{f}) - Quant^h(\mathcal{Z}^{h}) \right\|^2 + \\
\left\| Quant^b(\mathcal{Z}^{f}) - Quant^b(\mathcal{Z}^{b}) \right\|^2
\end{split}
\end{equation}
where \( Quant^h(\cdot) \) and \( Quant^b(\cdot) \) denote the quantization functions for the hand and body codebooks, respectively. 

The multimodal quantization loss aligns the integrated latent code $\mathcal{Z}^{f}$ with the learned motion code, a critical step since gesture synthesis is obtained through the quantized multimodal representation. Specifically, $\mathcal{Z}^{f}$ is vector-quantized using separate hand and body codebooks before being decoded by their respective VQ decoders. This process ensures that both hand and body movements contribute effectively to the final output. Formally, the generated gestures are given by:
\begin{equation}
\scriptsize
\hat{G} = \hat{G}^h \oplus \hat{G}^b = \mathcal{D}_{m}^h\bigl(Quant^h(\mathcal{Z}^{f})\bigr) \oplus \mathcal{D}_{m}^b\bigl(Quant^b(\mathcal{Z}^{f})\bigr),
\end{equation}
where $\oplus$ denotes concatenation, jointly synthesizing hand and body motions (\ie $\hat{G}$).
 
\subsubsection{Gesture Semantic Relevance Loss}
To prioritize the generation of semantically meaningful gestures (e.g., iconic, metaphoric, or deictic), which are less frequent than beat gestures, we introduce a semantic relevance loss. This loss emphasizes semantic annotations while preventing over-penalization of minor deviations. Formally, it is defined as:
\begin{equation}
\scriptsize
\label{eq:loss_local_semantic}
\mathcal{L}_{\text{semantic-relevance}} = \mathbb{E}\Bigl[\lambda\,\Psi\bigl(\mathbf{G} - \hat{\mathbf{G}}\bigr)\Bigr],
\end{equation}
where \(\lambda\) is the annotation relevance factor, and \(\Psi(\cdot)\) is a piecewise function that applies a quadratic penalty for small errors and a linear penalty for larger ones:
\begin{equation}
\scriptsize
\Psi\bigl(\mathbf{G} - \hat{\mathbf{G}}\bigr) = 
\begin{cases} 
\frac{1}{2}\bigl(\mathbf{G} - \hat{\mathbf{G}}\bigr)^2, & \text{if } \bigl|\mathbf{G} - \hat{\mathbf{G}}\bigr| < \alpha,\\[1ex]
\alpha\Bigl(\bigl|\mathbf{G} - \hat{\mathbf{G}}\bigr| - \frac{1}{2}\alpha\Bigr), & \text{otherwise},
\end{cases}
\end{equation}
with \(\alpha=0.01\).

\paragraph{Combined Objective Functions.}
Finally, the overall objective is:
\begin{align}
\scriptsize
    \begin{split}
\mathcal{L}_{\text{SemGes}}
=  \, \mathcal{L}_{\text{semantic-coherence}}
+
 \, \mathcal{L}_{\text{semantic-relevance}}
+ \, \mathcal{L}_{\text{quantization}},
\end{split}
\end{align}
which jointly optimizes the model to generate gestures that are semantically coherent at both global and fine-grained levels while remaining faithful to the Stage~1 motion prior.

\begin{algorithm}[t]
\caption{Long Gesture Sequence Algorithm}
\label{alg:long_seq_motion}
\begin{algorithmic}[1]
\Require Audio $\mathcal{A}$, aligned speech transcript $\mathcal{S}$, and speaker ID $\mathcal{I}$; Pre-trained codebooks and motion decoder (Stage~1)
\Ensure Long-sequence gesture $\hat{M}$
\State Partition $(\mathcal{A}, \mathcal{S}, \mathcal{I})$ into clips $\{(\mathcal{A}_c, \mathcal{S}_c, \mathcal{I}_c)\}_{c=1}^{C}$
\State Compute latent representation: $\mathcal{Z}^{f} \gets \mathrm{Encode}(\mathcal{A}, \mathcal{S}, \mathcal{I})$
\State Quantize: $\mathcal{Z}^{e} \gets \mathrm{VectorQuantize}(\mathcal{Z}^{f})$
\State Decode initial clip: $\hat{M}_1 \gets \mathrm{Dec}(\mathcal{Z}^{e})$
\For{each clip $c = 2$ to $C$}
    \State Set first 4 frames of $\hat{M}_c$ to the last 4 frames of $\hat{M}_{c-1}$
    \State Generate remaining frames of $\hat{M}_c$
\EndFor
\State \Return $\hat{M}$
\end{algorithmic}
\end{algorithm}

\subsection{Inference of Long Gesture Sequences}
Generating long sequences of gestures is challenging due to the need to maintain coherence and smooth transitions. Our Long-Sequence Gesture Motion algorithm (Alg.~\ref{alg:long_seq_motion}) addresses these challenges by partitioning the input speech, transcript, and speaker identity into aligned clips. For each clip, a multimodal latent representation is computed using our cross-modal encoder, vector-quantized via the Stage~1 codebooks, and decoded into gesture motions. Overlapping 4-frame segments between clips provides continuity, resulting in extended, naturalistic gesture sequences.

\section{Experimental Setup}
\label{sec:Experiments}
\subsection{Datasets}
\begin{table*}[ht]
\centering
\caption{Comparison of SemGesGen with other methods on the BEAT and TED-Expressive datasets. For BEAT, we compare with CaMN \cite{liu2022beat}, DiffGesture \cite{zhu2023taming}, LivelySpeaker \cite{zhi2023livelyspeaker}, and DiffSheg \cite{chen2024diffsheg}. The same methods are evaluated on TED-Expressive. SRGR is not applicable (denoted with --) for TED-Expressive as it does not contain annotations for semantic relevance of gestures.}
\label{tab:comparisions_with_other_models}
\resizebox{0.9\textwidth}{!}{%
\begin{tabular}{l c c c c || l c c c}
\toprule
\multicolumn{5}{c||}{\textbf{BEAT}} & \multicolumn{4}{c}{\textbf{TED-Expressive}} \\
\cmidrule(r){1-5} \cmidrule(l){6-9}
\textbf{Method} & \textbf{FGD $\downarrow$} & \textbf{BC $\uparrow$} & \textbf{Diversity $\uparrow$} & \textbf{SRGR $\uparrow$} & \textbf{Method} & \textbf{FGD $\downarrow$} & \textbf{BC $\uparrow$} & \textbf{Diversity $\uparrow$}\\
\midrule
CaMN \cite{liu2022beat} & 8.510 & 0.797 & 206.789 & 0.231 & CaMN \cite{liu2022beat} & 7.673 & 0.642 & 156.236  \\
DiffGesture \cite{zhu2023taming} & 9.632 & 0.876 & 210.678 & 0.106 & DiffGesture \cite{zhu2023taming} & 9.326 & 0.662 & 119.889  \\
LivelySpeaker \cite{zhi2023livelyspeaker} & 13.378 & 0.891 & 214.946 & 0.229 & LivelySpeaker \cite{zhi2023livelyspeaker} & 8.145 & 0.691 & 119.231  \\
DiffSheg \cite{chen2024diffsheg} & 6.623 & \textbf{0.922} & 257.674 & 0.250 & DiffSheg \cite{chen2024diffsheg} & 8.457 & \textbf{0.712} & 108.972  \\
\midrule
SemGes (Ours)  & \textbf{4.467} & 0.453 & \textbf{305.706} & \textbf{0.256} & SemGes (Ours)  & \textbf{7.263} & 0.671 & \textbf{302.772}  \\
\bottomrule
\end{tabular}%
}
\vspace{-0.2cm}
\end{table*}

Our proposed methodology is evaluated on two benchmarks, namely, BEAT~\cite{liu2022beat} and the TED expressive dataset~\cite{liu2022learning}.
\textbf{The BEAT dataset} consists of 76 hours of multimodal recordings, which include speech audio recordings, speech transcriptions, and, more importantly, motion data collected from 30 participants, leveraging Motion Capture (MOCAP) technology. The participants expressed emotions in eight distinct scenarios across four languages.
The motion data contains joint rotation angles, which were designed for consistency across varying body sizes.
\textbf{The TED Expressive dataset}~\cite{liu2022learning} is segmented from TED Talk videos into smaller shots based on scene boundaries. ~\citet{liu2022learning}  extracted each frame's 2D human pose using OpenPose BEAT~\cite{cao2017realtime}. Using these 2D pose priors, ExPose \cite{pavlakos2019expressive} was employed to annotate the 3D upper body keypoints, including 13 upper body joints and 30 finger joints. Both datasets' training and validation samples are divided into 34-frame clips.

\paragraph{Cross-Validation.}
We evaluate our approach on the BEAT dataset, following the protocol in~\cite{liu2022beat}, where the data is randomly split into a 19:2:2 ratio for training, validation, and testing. Similarly, for the TED Expressive dataset, we adapt the protocol in~\cite{liu2022learning}, using a random split of 8:1:1 for training, validation, and testing.

\paragraph{Implementation Details.}
The details of the model architectures and training are provided in Section 2 of the Supplementary Materials.

\subsection{State-of-the-Art Baselines}
We compare SemGes against a set of representative state-of-the-art models that focus on semantic-driven gesture generation. The selected models achieved strong performance on the BEAT and TED-Expressive datasets, making them suitable for a fair comparison with our method. 
The selected models are as follows:

\begin{enumerate}
     \item \textbf{Cascaded Motion Network(CaMN) \cite{liu2022beat}} is the current benchmark model for the BEAT dataset. CaMN is based on LSTMs and integrates multiple input modalities, including audio, text, facial expressions, and emotion. Additionally, like SemGes, it leverages semantic relevance annotations to enhance gesture generation.

    \item \textbf{DiffSHEG~\cite{chen2024diffsheg}} is a state-of-the-art diffusion-based model for real-time speech-driven holistic gesture generation. It is conditioned on noisy motion, audio, and speaker ID. DiffSHEG introduces a Fast Out-painting-based Partial Autoregressive Sampling method to efficiently generate arbitrary-length sequences in real time. 

    \item \textbf{LivelySpeaker~\cite{zhi2023livelyspeaker}} generates semantically and rhythmically aware co-speech gestures by leveraging an MLP-based diffusion model. The model conditions gesture generation on text, noised motion, speaker ID, and audio to enable text-driven gesture control while incorporating global semantics.

    \item \textbf{ DiffGes~\cite{zhu2023taming}} models the diffusion and denoising processes within the gesture domain, enabling the generation of high-fidelity, audio-driven gestures conditioned on both audio and gesture inputs. Several recent studies\cite{liang2022seeg,ao2023gesturediffuclip} have also demonstrated strong performance in this area.
\end{enumerate}
\noindent
We exclude certain models from our comparison. For instance, SEEG~\cite{liang2022seeg} and \cite{zhang2024semantic} rely on additional data annotations (e.g., Semantic Prompt Gallery or ChatGPT-generated annotations) that are not uniformly available
In addition, other works, such as \citet{ao2023gesturediffuclip,zhang2024semantic,pang2023bodyformer}, are excluded from our analysis due to the inaccessibility of their codebase. \citet{voss2023augmented} is omitted due to its high computational cost and the unavailability of annotations.\citet{ng2024audio,liu2024towards,mughal2024convofusion,yi2023generating,mughal2024retrieving,liu2025gesturelsm} are excluded as they primarily focus on holistic gestures with face and mesh data, which fall outside the scope of this work. 
Similarly,  \citet{qi2024weakly,chhatre2024emotional} are excluded, as their emphasis lies in emotion-driven gesture generation rather than the semantic aspects. Furthermore, \citet{ahuja2023continual, alexanderson2023listen,sun2023co,yang2023qpgesture,ye2022audio,habibie2022motion,liu2022learning,ahuja2022low} are omitted due to their lack of relevance to semantic-driven gesture generation.

\begin{table*}[ht]
\small
\centering
\caption{Ablation studies evaluating the contributions of key components in SemGes on the BEAT and TED-Expressive Datasets. For BEAT, performance is measured using FGD (lower is better), BC, Diversity, and SRGR, while for TED-Expressive, SRGR is not applicable (denoted as --).}
\label{tab:ablation}
\resizebox{0.9\textwidth}{!}{%
\begin{tabular}{l c c c c || l c c c}
\toprule
\multicolumn{5}{c||}{\textbf{BEAT}} & \multicolumn{4}{c}{\textbf{TED-Expressive}} \\
\cmidrule(r){1-5} \cmidrule(l){6-9}
\textbf{Model Variants} & \textbf{FGD $\downarrow$} & \textbf{BC $\uparrow$} & \textbf{Diversity $\uparrow$} & \textbf{SRGR $\uparrow$} & \textbf{Model Variants} & \textbf{FGD $\downarrow$} & \textbf{BC $\uparrow$} & \textbf{Diversity $\uparrow$} \\
\midrule
Baseline (VQVAE)                & 10.348 & 0.564 & 198.568 & 0.176 & Baseline (VQVAE)                & 10.682 & 0.612 & 114.692 \\
w/o Semantic Coherence Module   & 8.053  & 0.556 & 249.550 & 0.180 & w/o Semantic Coherence Module   & 7.924  & 0.623 & 109.256 \\
w/o Semantic Relevance Module    & 7.549  & \textbf{0.573} & 245.319 & 0.195 & w/o Semantic Relevance Module    & --     & --    & --\\
w/ SpeechCLIP Encoder           & 6.787  & 0.468 & 289.621 & 0.245 & w/ SpeechCLIP Encoder           & 7.341  & 0.605 & 245.680 \\
SemGes (Ours)                           & \textbf{4.467}  & 0.453 & \textbf{305.706} & \textbf{0.256} & SemGes (Ours)                           & \textbf{7.263}  & \textbf{0.671} & \textbf{302.772}\\
\bottomrule
\end{tabular}%
}
\vspace{-0.2cm}
\end{table*}
\section{Quantitive Objective Evaluations}
\paragraph{Evaluation Metrics.}
We employ four standard objective metrics for evaluating the quality of gesture generation, namely, Fr\'echet Gesture Distance (FGD) \cite{yoon2020speech}, Beat Consistency Score (BC) \cite{li2021ai}, Diversity~\cite{li2021audio2gestures}, and Semantic-Relevant Gesture Recall (SRGR)~\cite{liu2022beat}. 

\textbf{FGD} measures how the generated gestures resemble real motion distributions by embedding sequences into a latent space via a pre-trained autoencoder. In contrast, \textbf{BC} focuses on synchronization with speech, measuring the alignment between speech onsets (audio beats) and motion beats, which are identified as velocity minima in upper-body joints (excluding fingers). Meanwhile, \textbf{Diversity} captures the variability of generated motions by computing the average $L1$ distance between pairs of $N$ generated clips. Finally, \textbf{SRGR} assesses semantic relevance by determining how well generated gestures align with the annotated semantic gestures. Further details on the objective metrics are included in the Supplementary Materials (Section 1).

\paragraph{Comparisons with Other Models.} 
Table~\ref{tab:comparisions_with_other_models} compares the performance of our approach against four baseline methods across four evaluation metrics. As highlighted in the table, SemGes outperforms the baselines in FGD, Diversity, and SRGR.

For the BEAT dataset, our approach achieves the highest SRGR, which we attribute to the exploitation of semantic relevance information in our training objectives. In addition, our approach shows a significant improvement in FGD and Diversity, indicating a closer alignment with the ground truth gesture distribution and a broader range of generated gestures compared to the second-best baselines.
The performance on the Beat Consistency (BC) metric is lower for our method. This is expected given our focus on improving semantic awareness of the generated gestures rather than optimizing for strict temporal alignment between rhythmic beat gestures and speech. In addition, the BC metric can be sensitive to rapid, jittery movements; even minor motion artefacts may be mistakenly counted as additional beats, thereby increasing the BC score artificially- a phenomenon also observed in the diffusion-based baselines, as further illustrated in our supplementary video.

We evaluate how our model handles the trade-off between semantic and beat scores by testing the model on beat-dominant gestures (without semantic content). The results show a significantly higher Beat score (0.689) than the full dataset Beat score (0.453). This confirms rhythmic consistency in beat-focused contexts. We provide additional evaluation in the supplementary materials (Section 3) to show how the model handles difficult cases (such as noisy speech or misaligned speech).

Note that the TED Expressive dataset lacks annotations for gesture semantic relevance, so SRGR is not applicable, and the semantic relevance loss was omitted during training. Nevertheless, SemGes produces diverse, naturalistic gestures on TED Expressive, outperforming baselines in FGD and Diversity metrics.


\paragraph{Ablation Study.}
We evaluate the contributions of key components in SemGes through ablation experiments. First, we assess a baseline VQ-VAE model (Stage 1 only), which uses two stacked encoder-decoder blocks and an MLP. In this experiment, we test its ability to generate gestures,  conditioned on audio, masked motion, and speaker identity. As shown in Table~\ref{tab:ablation}, this baseline underperforms compared to state-of-the-art methods (Table~\ref{tab:comparisions_with_other_models}). As a result, we motivate our two-stage design where the VQ-VAE is reserved to learn the motion latent space and Stage 2 leverages speech and identity conditioning to generate gestures. 

Next, we examine Stage 2 by removing its components: (i) the Semantic Coherence Loss, (ii) the Semantic Relevance Loss, and (iii) by replacing the HuBERT-based speech encoder with SpeechCLIP. Results in Table~\ref{tab:ablation} show that removing either the Semantic Coherence or Relevance Loss degrades FGD, Diversity, and SRGR scores, highlighting their roles in aligning gesture representations with textual semantics and capturing semantic importance. In addition, replacing the speech encoder results in marginal gains. The semantic encoder is fixed as FastText, which we believe is sufficient to capture the necessary semantic information~\cite{athiwaratkun2018probabilistic}. Overall, these results confirm the importance of each module in generating semantics-aware gestures.

\begin{figure*}[th]
  \centering  \includegraphics[width=0.99\linewidth]{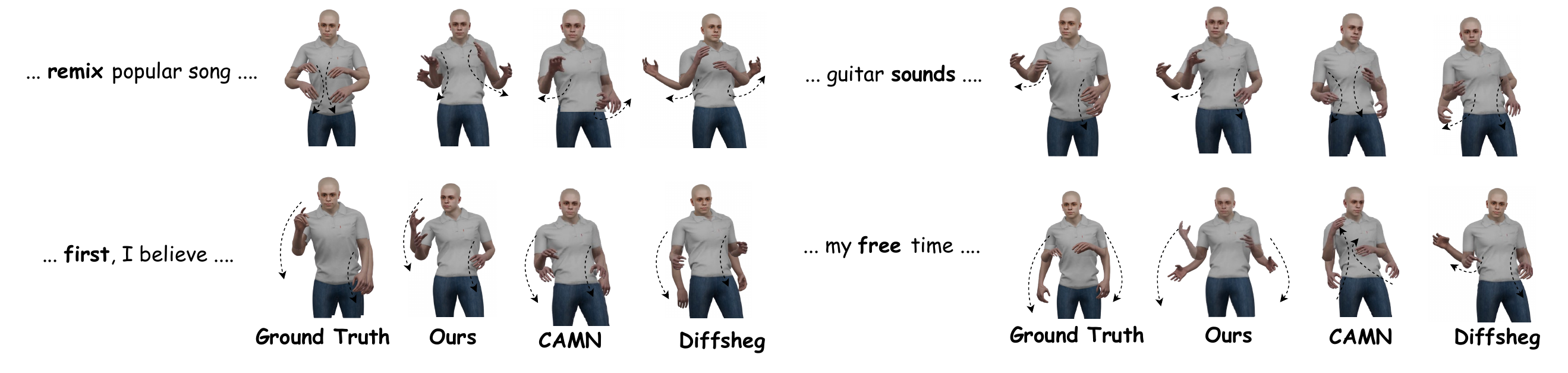}
    \caption{Comparisons with baselines and ground truth gestures. Compared to the baseline method, our approach generates gestures that are aligned with speech content (semantics). For instance, when the speaker says \textcolor{blue}{``remix''}, our method produces gestures where the character raises both hands to emphasize the word before gradually lowering them—a movement that other methods fail to achieve. Similarly, when uttering \textcolor{blue}{``first''}, our method generates a raised hand gesture, producing an \emph{iconic gesture}.}
    \label{fig:rendered_motion}
     \vspace{-0.2cm}
\end{figure*}
\section{Qualitative \& Subjective Evaluations}
\paragraph{Visualization Comparisons.}
Before presenting the subjective ratings of the generated gestures, Figure~\ref{fig:rendered_motion} provides a visual comparison between the ground truth, results from our approach and two baseline models. We use examples from the BEAT dataset. It is clear from the figure that our approach not only achieves better speech-gesture alignment but also produces gestures that are more naturalistic, diverse, and semantically aware. 
For example, while CaMN generates smooth movements, its gestures tend to be slower and less varied compared to our model. Additionally, the baseline methods show varying degrees of jitter—DiffGesture shows the highest jitter, followed by LivelySpeaker and DiffSheg, with CaMN displaying the least. Although CaMN includes semantic information, our approach strikes a more effective balance, generating gestures that align with actual motion, as shown also with the objective metrics. 
Based on these qualitative observations, our subsequent rating study focuses on evaluating gestures produced by the ground truth, our model, CaMN, and DiffSHEG.

\paragraph{User Ratings of Generated Gestures.}
We conducted a user study using 40-second video clips from the BEAT test set, featuring subjects narrating six topics. Thirty native English speakers from the United Kingdom and the United States participated, with an average age of $36\pm20$ years and a female-to-male ratio of approximately 2:1. Each participant evaluated 24 videos generated by the ground truth, CaMN, DiffSHEG, and our model over a study duration of that lasted on average $27\pm5$ minutes. For data quality, participants were required to pass attention verification questions, \ie, correctly answering at least two out of four questions regarding the narration topic. Participants rated the videos on three criteria: naturalness, diversity, and alignment with speech content and timing on a scale from 1 to 5. The videos were presented in a randomized order to avoid bias. In Section 3 of the Supplementary Materials, we provide screenshots and more details on the user study and interface.

\begin{figure}[htbp]
  \centering
  \includegraphics[width=1.0\linewidth]{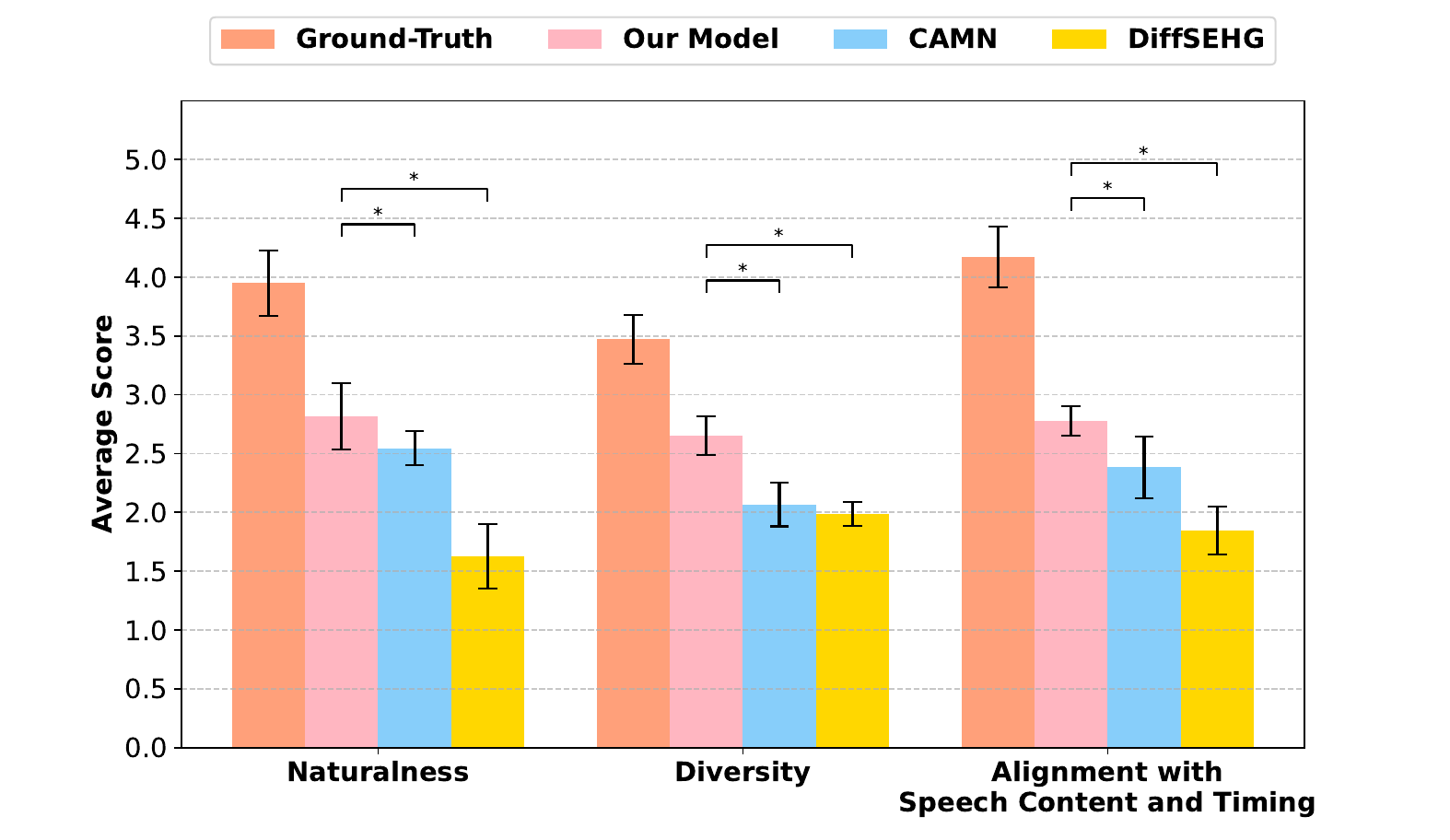}
  \caption{Average ratings of users for ground truth gestures and gestures generated through our approach, CAMN, and DiffSHEG. The bars illustrate the average user ratings across three metrics: naturalness, diversity, and alignment with speech content and timing. Statistical t-tests show that our approach received significantly higher ratings than CAMN and DiffSHEG, with $p < 0.05$.}
  \label{fig:user_ratings}
  \vspace{-0.2cm}
\end{figure}

Figure~\ref{fig:user_ratings} shows that ground-truth gestures received an average rating of 4 across all metrics, establishing an upper bound and validating the participant survey. Our model received the highest ratings among the generated gestures, significantly outperforming CaMN and DiffSHEG in naturalness, synchronization, and diversity (indicated by the $\*$ in Figure~\ref{fig:user_ratings}). These results prove that our approach produces gestures that are more natural, better aligned with speech, and more diverse than those generated by SOTA baselines.

\section{Conclusion}
\label{sec:Conclusion}
We proposed SemGes, a novel two-stage approach to semantic grounding in co-speech gesture generation by integrating semantic information at both fine-grained and global levels. In the first stage, a motion prior generation module is trained using a vector-quantized variational autoencoder to produce realistic and smooth gesture motions. Building upon this model, the second stage generates gestures from speech, text-based semantics, and speaker identity while maintaining consistency between gesture semantics and co-occurring speech through semantic coherence and relevance modules. Subjective and objective evaluations show that our work achieves state-of-the-art performance across two public benchmarks, generating semantics-aware and diverse gestures. Future direction and limitations are discussed in Section 5 of the Supplementary Materials. 

\section*{Acknowledgement}
\label{sec:Acknowledgement}
The project is funded by the Max Planck Society.
We thank Sachit Misra for his invaluable assistance with rendering Avatar characters. We extend our gratitude to the members of the Multimodal Language Department at Max Planck Institute for Psycholinguistics for their feedback.
{
    \small
    \bibliographystyle{ieeenat_fullname}
    \bibliography{main}
}
\end{document}